\documentclass[runningheads]{llncs}

\usepackage[utf8]{inputenc} 
\usepackage[T1]{fontenc}    
\usepackage{hyperref}       
\usepackage{url}            
\usepackage{booktabs}       
\usepackage{amsfonts}       
\usepackage{nicefrac}       
\usepackage{microtype}  
\usepackage{graphicx}
\usepackage{multirow}
\usepackage{makecell}
\usepackage{tabularx}
\usepackage{textcomp}
\usepackage{enumitem}
\usepackage{xcolor}
\usepackage{float}
\usepackage[htt]{hyphenat}
\usepackage[square,numbers]{natbib}
\begin{document}
\title{Evaluating Polish linguistic and cultural competency in large language models}
\titlerunning{Evaluating Polish linguistic and cultural competency in LLMs}

\author{Sławomir Dadas \and Małgorzata Grębowiec \and Michał Perełkiewicz \and Rafał Poświata}
\authorrunning{S. Dadas, M. Grębowiec, M. Perełkiewicz, R. Poświata}
\institute{National Information Processing Institute, Warsaw, Poland\\
\email{\{sdadas,mgrebowiec,mperelkiewicz,rposwiata\}@opi.org.pl}}

\maketitle 
\begin{abstract}
Large language models (LLMs) are becoming increasingly proficient in processing and generating multilingual texts, which allows them to address real-world problems more effectively. However, language understanding is a far more complex issue that goes beyond simple text analysis. It requires familiarity with cultural context, including references to everyday life, historical events, traditions, folklore, literature, and pop culture. A lack of such knowledge can lead to misinterpretations and subtle, hard-to-detect errors. To examine language models' knowledge of the Polish cultural context, we introduce the Polish linguistic and cultural competency benchmark, consisting of 600 manually crafted questions. The benchmark is divided into six categories: history, geography, culture \& tradition, art \& entertainment, grammar, and vocabulary. As part of our study, we conduct an extensive evaluation involving over 30 open-weight and commercial LLMs. Our experiments provide a new perspective on Polish competencies in language models, moving past traditional natural language processing tasks and general knowledge assessment.

\end{abstract}

\section{Introduction}
Recently, we have observed a dynamic development of large language models (LLMs). This led to profound changes in the approach to natural language processing. As a result, small models fine-tuned to perform specific tasks are gradually being replaced by large generative models, trained on massive text corpora and millions of instructions covering a wide variety of problems. From a business and end-user perspective, this simplifies the practical use of language models, since prompt engineering and few-shot learning are techniques that do not require specialized knowledge, as opposed to earlier approaches that involved re-training the models for each task. However, evaluating modern language models poses a significant challenge. Firstly, their ability to solve various types of tasks, and even generalize to tasks not seen during training, requires a multifaceted approach to evaluation. Secondly, these models generate responses in plain text, allowing a high degree of freedom in terms of response style and formatting.   Assessing the correctness of a textual response is more challenging than evaluating a structured response such as a numerical value, class label, probability, or answer from a closed vocabulary.

In recent years, various approaches to evaluating LLMs have been proposed. At the same time, there is a growing awareness of their drawbacks and limitations \citep{10.1145/3641289, mcintosh2024inadequacies, laskar2024systematicsurveycriticalreview}. Therefore, it is crucial to use a combination of several benchmarks to verify various aspects of understanding and generation of natural language. One such aspect is multilinguality. Currently, an increasing number of models declare multilingual capabilities. Initially, these were mainly commercial models such as GPT-4 \citep{openai2024gpt4technicalreport}, Claude \citep{TheC3} or Gemini \citep{reid2024gemini}, but the trend has also extended to open-weights models. This publication focuses on the Polish language, which is often included among the supported languages. There are families of models that mention Polish directly (Qwen2 \citep{yang2024qwen2}, Command-R \citep{cohereIntroducingCommand2024}, Aya \citep{aryabumi2024aya}) and those that declare broad support for multilinguality and are known to work well for Polish (Mistral \citep{jiang2023mistral, jiang2024mixtral}, Llama 3.1+ \citep{dubey2024llama}). It is also worth mentioning initiatives focusing solely on the Polish language such as SpeakLeash\footnote{\url{https://speakleash.org/}} or PLLuM \citep{kazienko2024}.

\begin{figure}
  \centering
  \includegraphics[scale=0.74]{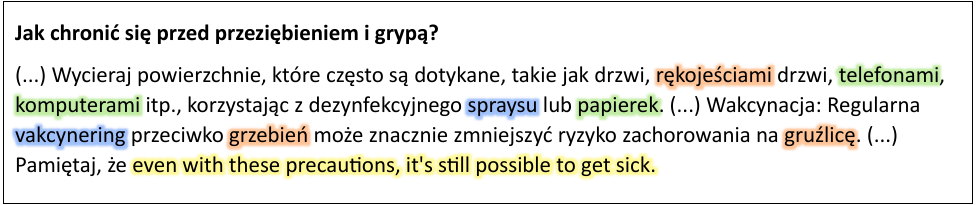}
  \caption{Example showing various types of errors made by a multilingual model that was undertrained on Polish data. The user prompt is: \emph{How do you protect yourself from cold and flu?} (\emph{Jak chronić się przed przeziębieniem i grypą?}). In the model's response, we can observe the following errors: a) \textbf{morphological} (green) - using the wrong form of a word; b) \textbf{lexical} (orange) - using a word that does not fit the context; c) \textbf{word-formation} (blue) - attempt to create a new word by imitating a word in another language, usually English; d) \textbf{changing the language} of the answer completely (yellow). The example includes snippets of the response generated by the Qwen2.5 7B model, using the interface provided by the model's authors. For larger Qwen2.5 models the frequency of such errors decreases.}
  \label{fig:example}
\end{figure}

In practice, the level of Polish proficiency among multilingual models leaves much to be desired. Models with a small number of parameters can struggle to generate coherent and grammatically correct text. As an example, Figure \ref{fig:example} shows the errors made by the Qwen2.5 model \citep{yang2024qwen2} when prompted to answer a simple question in Polish. For larger models, such problems are less frequent. However, the underrepresentation of Polish texts in the pre-training corpus can result in the model lacking Polish cultural context, which is essential for language understanding. The language includes references to historical events, folklore, literature, or pop culture can carry specific meanings in a particular culture but may not be understood or may have different associations in another culture \citep{Stan_2023}. As a result, models that on the surface may perform well in understanding and generating language will have difficulties correctly interpreting texts containing such references.

\subsection{Contributions}
The goal of this publication is to examine both the linguistic and cultural competence of LLMs in Polish. To this end, we introduce a benchmark consisting of 600 manually prepared questions. The questions are divided into six categories representing different aspects of knowledge about Poland and the Polish language: history, geography, culture \& tradition, art \& entertainment, grammar, and vocabulary. Apart from thematic diversity, we also ensured diversity in the form of questions. In addition to questions that require a direct answer, the benchmark includes questions that involve combining facts, aggregating knowledge about different concepts, and sorting or filtering concepts according to specified characteristics. The evaluation of answers is deterministic and involves a rule-based system, inspired by benchmarks such as IFEval \citep{zhou2023instruction} or BBH \citep{suzgun2022challenging}. For each question, we define a set of rules that must be met for the question to be considered as passed. For closed-ended questions, these rules are explicit and easy to define. For open-ended questions,  the presence or order of certain words, phrases, or regular expressions is checked. As part of this study, we evaluated over 30 open-weights and commercial LLMs. We also launched a demo website\footnote{\url{https://huggingface.co/spaces/sdadas/plcc}} featuring an up-to-date leaderboard,  examples of questions from each category, and a description of our methodology.

\subsection{Related work}
Most methods for evaluating large language models can be divided into four categories:
\begin{itemize}[wide,labelwidth=0pt,labelindent=0pt,itemsep=0pt,topsep=5pt]
\item[$\bullet$]{\textbf{Closed-ended questions and structured answers} - Benchmarks of this type require the formulation of tasks for the LLM in such a way that the answer follows a format imposed by the authors so that it can be verified deterministically. The most common type of such tasks are single- or multiple-choice questions, in which the model is presented with answers from which it must select the correct ones. Among these types of tests, the popular ones include GPQA \citep{rein2024gpqa} and different variants of MMLU \citep{hendryckstest2021,wang2024mmlu}. In addition, the group also includes tests that expect a structured answer, for example, datasets of math questions such as MATH \citep{hendrycksmath2021} or GSM8K \citep{cobbe2021gsm8k}. The category also incorporates many classic NLP tasks such as classification, regression, or information extraction.}
\item[$\bullet$]{\textbf{LLM-as-judge} - In this approach, a different language model is used to assess model's performance, usually a larger and more competent or trained specifically for the purpose of evaluation. Popular examples of such benchmarks are MT-BENCH \citep{10.5555/3666122.3668142} and AlpacaEval \citep{dubois2024lengthcontrolled}. Their advantage is the ability to evaluate models in tasks requiring free-form and long answers. On the other hand, the main disadvantage is the dependence on an external judge, which may be biased from the data on which it was trained. Consequently, results on such benchmarks may vary depending on which model is used as a judge. In addition, the cost of LLM-as-judge evaluation is higher than that of deterministic methods.}
\item[$\bullet$]{\textbf{Human evaluation} - An evaluation method that uses human annotators to rate the responses of LLM directly. This is the most costly of the described methods, as it requires manual labor. However, it allows for examining unique aspects of the models' performance, that is, the alignment of responses with human preferences, which is not possible with automated benchmarks. A notable example of such a benchmark is Chatbot Arena \citep{chiang2024chatbot}, a crowdsourced platform in which users rank pairs of answers from randomly selected models.}
\item[$\bullet$]{\textbf{Rule-based and programmatic evaluation} - This category includes methods that verify the correctness of the model's responses based on programmed rules. This approach can only be applied to specific types of problems. In addition, each type of task requires a set of individually crafted rules or hard-coded logic responsible for evaluation. For example, the ability of models to generate code can be verified by using an external compiler or interpreter, enabling to run the code and compare the output with the expected result. This type of approach is used by benchmarks such as HumanEval \citep{chen2021evaluating} or BigCodeBench \citep{zhuo2024bigcodebench}. Even more sophisticated evaluation frameworks are used in LLM-based agent systems. For example, AgentBench \citep{liu2024agentbench} allows models to interact with external software such as an operating system or database and then examines the agent's ability to perform specific actions using that software. Text generation tasks typically use simpler rules that check the content of the answer itself or additional conditions that have been imposed on the model in the prompt. Such rules are used by the IFEval \citep{zhou2023instruction}, BBH \citep{suzgun2022challenging} or MuSR \citep{sprague2024musr} benchmarks, among others.}
\end{itemize}

For the Polish language, there already exist benchmarks created to evaluate LLMs. Several of them were developed as part of the work on the Bielik model \citep{ociepa2024bielik7bv01polish}. These include adaptations of existing English benchmarks to Polish, such as MT-BENCH, EQ-BENCH, and Chatbot Arena\footnote{\url{https://huggingface.co/speakleash}}. In addition to these, there are the Open PL LLM Leaderboard and the Polish Medical Leaderboard, which do not introduce new datasets, instead aggregating pre-existing ones. The first aggregates mostly classification tasks from older NLP benchmarks \citep{rybak2020klej,dadas2020evaluation} and a small number of simple generative Q\&A tasks \citep{rybak2024polqa,10.1145/3587259.3627548}. The second contains closed-ended questions from the official medical and specialty exams \citep{grzybowski2024polishmedicalexamsnew}.

To the best of our knowledge, no dataset yet focuses on evaluating Polish linguistic and cultural competence in LLMs. Recently, such benchmarks for other languages have begun to appear, including multilingual ones covering groups of nations and cultures \citep{myung2024blend,chiu2024culturalbenchrobustdiversechallenging,etxaniz2024bertaqalanguagemodelsknow,mousi2024aradicebenchmarksdialectalcultural}. This suggests that this is an important issue that, with the popularization of LLMs, is starting to attract the interest of researchers.

\section{The benchmark}
We present the Polish linguistic and cultural competency benchmark, comprising hand-crafted questions designed to evaluate LLM's factual knowledge of Polish culture, tradition, and language. The level of difficulty of the questions varies, from those that would be answered by the majority of Poles to detailed questions focusing on region-specific culture or ethnic minorities. The questions have been phrased in such a way that it is deterministically possible to verify their correctness. Approximately half of them are various forms of closed-ended questions such as single-choice, multiple-choice, matching two sets of concepts, or filtering concepts from a list. The rest of the questions usually require answers consisting of a single sentence containing a specific fact or a set of facts. Open-ended questions allow some freedom for the model to generate an answer, but the response should refer to specific entities such as people, dates, numbers, places, certain concepts or phrases. In addition, each question is typically supplemented with instructions for the model, imposing a specific form of answer - short, precise, without additional comments or elaborate explanations. The dataset has been divided into six categories, with 100 questions in each. These categories include:

\begin{itemize}[wide,labelwidth=0pt,labelindent=0pt,itemsep=0pt,topsep=5pt]
\item[$\bullet$]{\textbf{Culture \& tradition} - The category contains questions about beliefs and religion, which are derived from both Christian and folk traditions, as well as Slavic mythology. It also contains questions about pop culture, including characters and events that have had a significant impact on Polish society. Finally, we also included questions about everyday life, covering Polish customs, cuisine, and clothing, among other topics.}
\item[$\bullet$]{\textbf{Art \& entertainment} - This category focuses on fields of art such as literature, painting, sculpture, theater, music, dance, or film. The questions cover works and figures related to Polish art. In addition, this category also features questions related to entertainment, including sports, popular music, television, radio, and people associated with show business.}
\item[$\bullet$]{\textbf{Geography} - The category covers Polish geography and is divided into four subcategories. Two of them deal with structures and phenomena of natural (e.g., mountains, rivers, climate) and man-made origin (e.g., cities, tourist attractions, industrial plants, mines). The third subcategory deals with socio-political geography, which includes questions on population, social problems, national borders, and administrative units. The last group is biogeography, which covers the fauna and flora of Poland.}
\item[$\bullet$]{\textbf{History} - The category covers Polish history from the time of Mieszko I to the present day. It is divided into subcategories corresponding to historical periods. It contains questions about important historical events and figures of significance to Polish politics, science, and humanities.}
\item[$\bullet$]{\textbf{Grammar} - The category deals with the rules and principles that govern the structure of sentences in the Polish language, as well as the rules of spelling (orthography). The questions address both theoretical foundations and practical applications of grammar. In addition to orthography, the questions cover such topics as morphology, parts of a sentence, parts of speech, phonetics, word formation, or rhetorical figures.}
\item[$\bullet$]{\textbf{Vocabulary} - The category verifies LLMs' ability to understand the meaning of words, idioms, sayings and proverbs. The questions mainly focus on less frequently used words and phrases. In addition, slang expressions, regionalisms, dialects, colloquial language, and youth language were also included. Apart from modern Polish, several questions also deal with archaisms and old language.}
\end{itemize}

\begin{figure}
  \centering
  \includegraphics[scale=0.52]{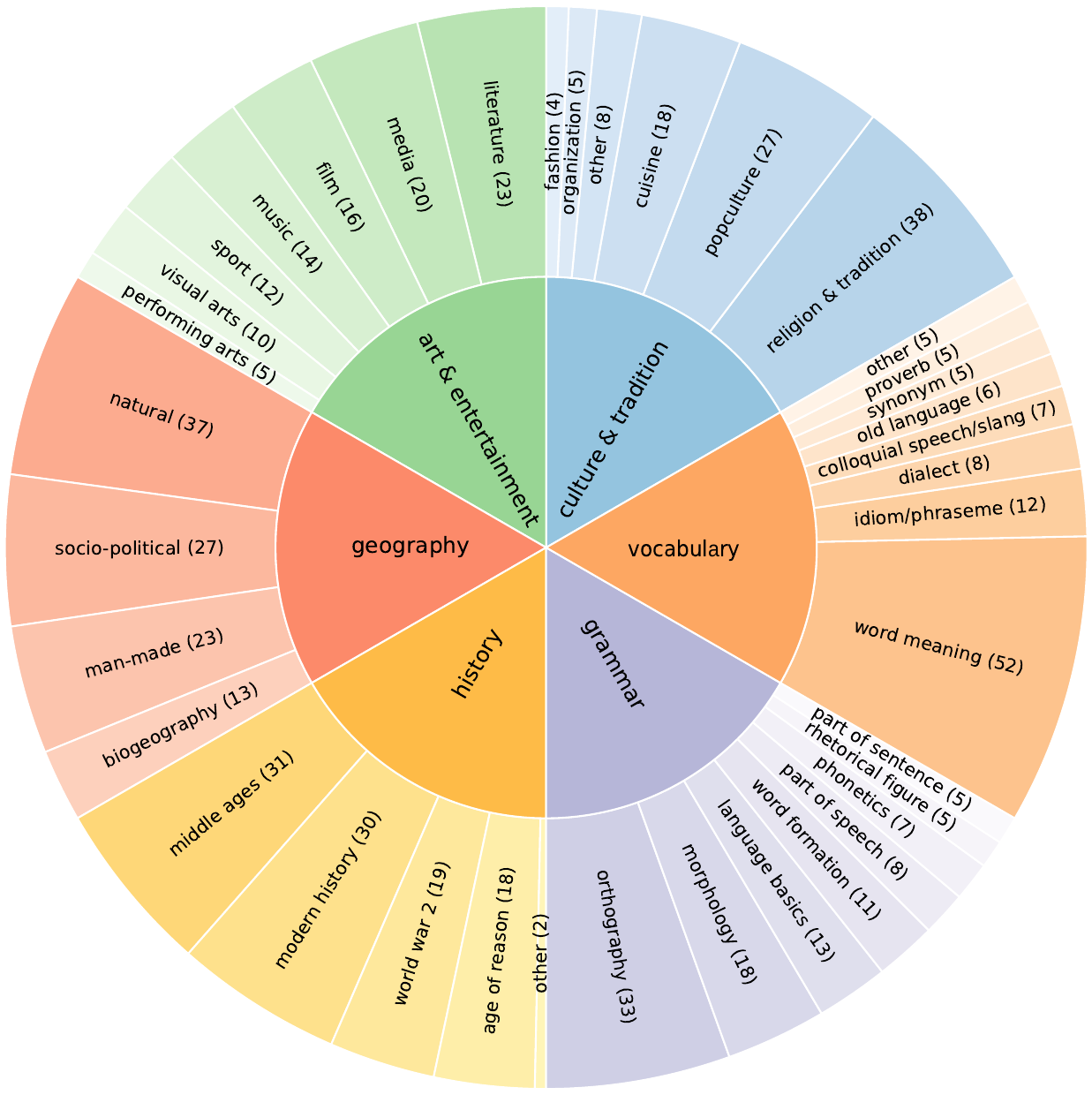}
  \caption{Distribution of questions by category and subcategory in our benchmark.}
  \label{fig:pie}
\end{figure}

\noindent{Figure \ref{fig:pie} shows a visualization of categories and subcategories in our dataset.}

\subsection{Grading}
Each question in the benchmark defines one or more conditions that must be met for an answer to be accepted. The scores are binary, a model can receive one or zero points for an answer. Partial points are not possible. This means that an answer is considered correct if and only if all the conditions defined in the question are satisfied. If at least one condition is not met, the model gets zero points for that question. The final score in the benchmark is calculated as a percentage of correct answers. 

The process of verifying a single question starts with sending it to the model and obtaining the answer. We do not use the system prompt in the evaluation, and the question is encoded using a chat template specific to each model as a single message with the user role. In addition, to ensure deterministic responses, the generation temperature parameter is set to 0. In the second step, we normalize the response. Normalization involves removing all characters except letters and numbers, making all letters lowercase, and then lemmatizing the text, that is, reducing all words to their base forms. Such normalization is necessary for languages with rich morphology like Polish, because only then we are able to match different forms of the same word between the model's response and the question's conditions. After normalization, we check each of the conditions defined in the question. The model scores a point only if all conditions have been verified successfully. Our benchmark supports the following condition types:
\begin{itemize}[wide,labelwidth=0pt,labelindent=0pt,itemsep=0pt,topsep=5pt]
\item[$\bullet$]{\textbf{Include} - Checks whether the words or phrases defined in the condition occurred in the model's response. In the simplest case, we can provide a list of comma-delimited expressions, and the condition will be satisfied only if all these expressions are found in the text. In practice, however, there may be more than one way to formulate the correct answer, so it is allowed to define alternative expressions for each item. In such a case, only one of the provided expressions needs to be in the answer. In other words, the \textbf{include} condition can verify any logical formula in conjunctive normal form (CNF), in which individual clauses check the occurrence of a word or phrase in the model's response. Moreover, the condition can be parameterized. Instead of the default behavior of matching all defined expressions, we can specify the minimum (\emph{include\_min}) and maximum (\emph{include\_max}) number of expressions that should be included in the response. In special cases, we can also disable lemmatization (\emph{lemmatize}) if we want to verify the occurrence of a specific, and not just any, form of a word.}
\item[$\bullet$]{\textbf{Exclude} - This is the inverse of the previous condition. It checks that none of the given words or phrases occur in the answer. The condition is verified in the similar way to \textbf{include}. We can also use the \emph{lemmatize} parameter to disable lemmatization when matching words.}
\item[$\bullet$]{\textbf{Order} - Checks whether the words or phrases defined in the condition occur in the response in the expected order. Verification of the condition is almost identical to \textbf{include}. The difference is that the positions of the first occurrence of the expression, or any of the alternative expressions, are saved. The condition is considered satisfied if and only if the positions are in ascending order. The condition can be used to verify questions involving sorting or matching information. The \emph{lemmatize} parameter can also be applied.}
\item[$\bullet$]{\textbf{Regex} - This is the most complex condition, which checks whether a defined regular expression occurs in the answer. If no additional parameters are specified, the condition is considered to be satisfied if at least one string matches the regular expression in the text. However, it is possible to control the acceptability criteria through parameters. For example, we can define the minimum (\emph{regex\_min}) and maximum (\emph{regex\_max}) number of occurrences allowed. Furthermore, it is possible to force all occurrences to have the same number of characters (\emph{regex\_match\_length}). Finally, we can introduce an additional dictionary criterion (\emph{regex\_match\_word}). If the parameter is set to \emph{true}, each matched string must be a valid word in Polish, which is verified using an external dictionary. Unlike the other conditions, \textbf{regex} is always verified on the unnormalized version of the response, so the \emph{lemmatize} parameter is not applicable.}
\end{itemize}

\section{Evaluation}
In this chapter, we present the results of LLMs evaluation on our Polish linguistic and cultural competency benchmark. In our experiments, we tested more than 30 commercial and open-weight models. A summary of the results is shown in Table \ref{tab:eval}. We report the percentage of correct answers by category and for all questions combined. The table is divided into sections grouping together models from the same provider. The order of the sections corresponds to the score of the best model in the group. For better readability, we have omitted multiple versions of the same model. In the case of models that are available in several versions such as the GPT-4o or the Claude Sonnet, the reported results refer to the latest available version as of the end of 2024. The best score in each column is bolded and underlined. It is worth noting the low cost of the evaluation, which is less than \$1 for the GPT-4o model according to the current pricing, allowing for a cost-effective comparison of multiple models. Based on the results, we can draw the following conclusions:

\noindent \textbf{Open vs. Commercial LLMs} - English-centric benchmarks suggest that the gap between the capabilities of open and commercial models is closing. In fact, from observing many popular LLM leaderboards, one can conclude that the best open models such as Llama 3.1 perform nearly as well as GPT-4o or Gemini-1.5-Pro. In contrast, our evaluation shows that for multilingual abilities, particularly understanding cultural and linguistic context, open models are still significantly behind those offered as a service. In our experiments, Gemini-Exp-1206 (83\%) proved to be the best model, followed closely by Claude Sonnet 3.5 and GPT-4o. These were the only three models that passed the 80\% mark. By comparison, the best-performing open-weight model scored approximately 12 points lower.

\noindent \textbf{The effectiveness of training on Polish data} - Bielik-2.3 is the only fully Polish model in the comparison, i.e. pre-trained on a Polish corpus and fine-tuned on Polish instructions. Despite its small size of just 11B parameters, it reached second place among open models, ranking only behind DeepSeek-V3. This example shows how much can be gained from building language-specific models. Bielik is based on Mistral-7b, which was upscaled and then continuously trained on Polish data. If we compare both models, we can see that Bielik improved by more than 40 points over Mistral.

\begin{table}[h]
  \centering
  \caption{Results of commercial and open-weight models on our benchmark. The table presents the partial scores for the six categories and the total score for all questions. For models that come in several versions, we used the latest available version for the evaluation as of the end of 2024. The best score in each column is shown in bold.}
  \aboverulesep=0ex
  \belowrulesep=0ex
  \setlength{\tabcolsep}{3pt}
  \renewcommand{\arraystretch}{0.8}
  \begin{tabular}{|rll|cccccc|c|}
    \toprule
    \textbf{Provider} & \textbf{Model} & \textbf{Size} & \scalebox{0.9}[1.0]{\textbf{Art}} & \scalebox{0.9}[1.0]{\textbf{Cul.}} & \scalebox{0.9}[1.0]{\textbf{Geo.}} & \scalebox{0.9}[1.0]{\textbf{Gra.}} & \scalebox{0.9}[1.0]{\textbf{His.}} & \scalebox{0.9}[1.0]{\textbf{Voc.}} & \textbf{Total} \\
    \toprule
    \textbf{Alibaba} & Qwen-2.5-72b & 72B & 25 & 30 & 45 & 45 & 54 & 36 & 39.17 \\
    \textbf{Alibaba} & Qwen-2.5-7b & 7B & 5 & 11 & 17 & 29 & 23 & 21 & 17.67 \\
    \hline
    \textbf{Cohere} & Command-R+ & 104B & 44 & 49 & 61 & 43 & 61 & 43 & 50.17 \\
    \textbf{Cohere} & Command-R7B & 8B & 14 & 18 & 33 & 23 & 27 & 22 & 22.83 \\
    \hline
    \textbf{Mistral} & Mistral-Large-2411 & 123B & 39 & 52 & 61 & 54 & 64 & 42 & 52.00 \\
    \textbf{Mistral} & Mixtral-8x22b & 141B & 45 & 41 & 59 & 50 & 69 & 35 & 49.83 \\
    \textbf{Mistral} & Mistral-Nemo & 12B & 20 & 13 & 26 & 31 & 28 & 20 & 23.00 \\
    \textbf{Mistral} & Mistral-7b-v0.3 & 7B & 22 & 9 & 27 & 27 & 30 & 16 & 21.83 \\  
    \textbf{Mistral} & Ministral-8b & 8B & 14 & 12 & 19 & 24 & 33 & 22 & 20.67 \\
    \hline
    \textbf{Meta} & Llama-3.1-405b & 405B & 56 & 57 & 74 & 57 & 73 & 43 & 60.00 \\
    \textbf{Meta} & Llama-3.3-70b & 70B & 43 & 40 & 59 & 49 & 65 & 37 & 48.83 \\
    \textbf{Meta} & Llama-3.1-8b & 8B & 19 & 13 & 31 & 29 & 25 & 19 & 22.67 \\
    \hline
    \textbf{SpeakLeash} & Bielik-2.3 & 11B & 58 & 61 & 68 & 49 & 76 & 61 & 62.17 \\
    \textbf{SpeakLeash} & Bielik-0.1 & 7B & 43 & 52 & 61 & 29 & 58 & 37 & 46.67 \\
    \hline
    \textbf{xAI} & Grok-2-1212 & - & 57 & 67 & 77 & 64 & 74 & 57 & 66.00 \\
    \hline
    \textbf{DeepSeek} & DeepSeek-V3 & 685B & 61 & 73 & 79 & 62 & 77 & 63 & 69.17 \\
    \hline
    \textbf{OpenAI} & GPT-4o & - & 82 & 89 & \textbf{\underline{86}} & 67 & 84 & 80 & 81.33 \\
    \textbf{OpenAI} & GPT-4-turbo & - & 61 & 74 & 79 & 56 & 76 & 56 & 67.00 \\
    \textbf{OpenAI} & GPT-4 & - & 49 & 63 & 67 & 58 & 72 & 48 & 59.50 \\
    \textbf{OpenAI} & GPT-4o-mini & - & 42 & 57 & 69 & 55 & 67 & 51 & 56.83 \\
    \textbf{OpenAI} & GPT-3.5-turbo & - & 39 & 38 & 55 & 41 & 51 & 36 & 43.33 \\
    \hline
    \textbf{Anthropic} & Claude-3.5-Sonnet & - & 77 & 87 & 85 & \textbf{\underline{79}} & \textbf{\underline{91}} & 77 & 82.67 \\
    \textbf{Anthropic} & Claude-3-Opus & - & 73 & 76 & 80 & 66 & 86 & 62 & 73.83 \\
    \textbf{Anthropic} & Claude-3.5-Haiku & - & 43 & 62 & 72 & 57 & 61 & 52 & 57.83 \\
    \hline
    \textbf{Google} & Gemini-Exp-1206 & - & \textbf{\underline{83}} & \textbf{\underline{90}} & \textbf{\underline{86}} & 69 & 88 & \textbf{\underline{82}} & \textbf{\underline{83.00}} \\
    \textbf{Google} & Gemini-2.0-Flash & - & 68 & 78 & 79 & 65 & 83 & 72 & 74.17 \\
    \textbf{Google} & Gemma-2-27b & 27B & 32 & 41 & 47 & 46 & 53 & 37 & 42.67 \\
    \textbf{Google} & Gemma-2-9b & 9B & 19 & 23 & 30 & 38 & 35 & 30 & 29.17 \\
    \hline
    \bottomrule
  \end{tabular}
  \label{tab:eval}
\end{table}

\noindent \textbf{Comparison with other Polish benchmarks} - In this publication, we explore aspects of LLMs that have so far been overlooked in Polish research. Unfortunately, current benchmarks for Polish mainly focus on general knowledge and classic NLP problems such as classification, often including tasks analogous to, or direct translations of English datasets. This means that they do little to test the models' actual ability to use and understand the language. Consequently, the results obtained by us differ significantly from previous evaluations. For example, models from the Qwen-2.5 family are usually highly rated on other Polish benchmarks mentioned in the related work section, meanwhile, in our experiments, they performed below expectations.

\noindent \textbf{Results in relation to model size} - We can observe significant differences in performance between small and large language models. There is a considerable drop in quality for smaller models, Bielik discussed earlier being the only exception. This can be explained by the fact that we are testing the models on a knowledge-based task. A higher number of parameters means a higher information density, which can be encoded in the model's weights, and therefore better performance for such tasks. In addition, the evaluation is mostly carried out on multilingual models, which are required to know facts about a variety of languages and cultures. It is understandable that a loss of information will be observed when we try to compress this knowledge into a small-sized model.

\begin{figure}
  \centering
  \includegraphics[scale=0.30]{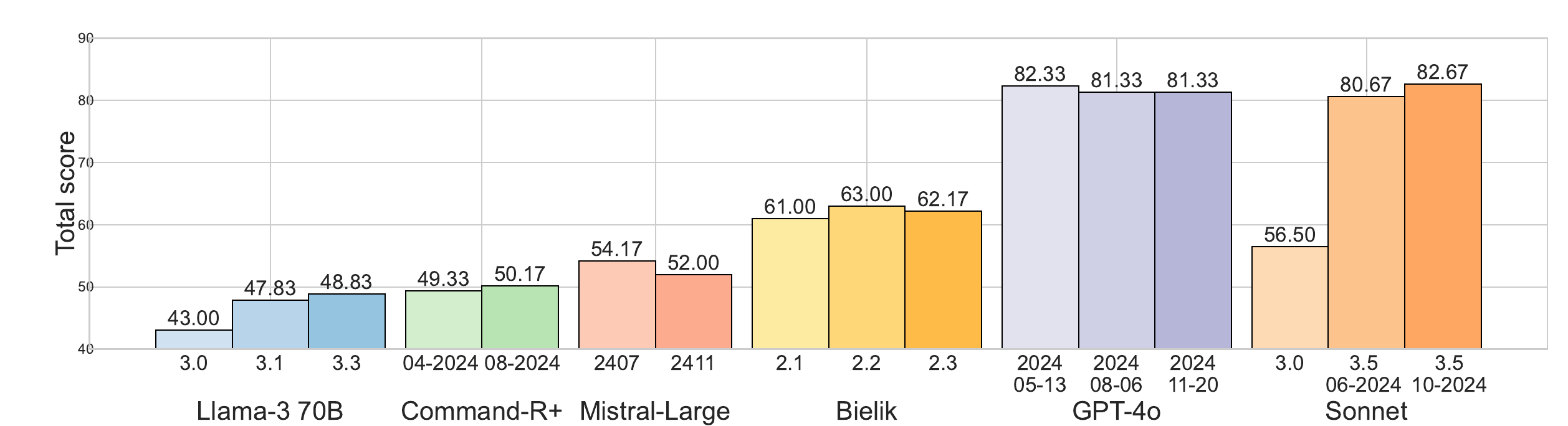}
  \caption{Comparison of benchmark scores for different versions of the same model.}
  \label{fig:versions}
\end{figure}

\noindent \textbf{Comparison with older models} - We performed an additional experiment in which we measured how the performance on our benchmark changed for different versions of several models. The results are shown in Figure \ref{fig:versions}. All models included in the experiment were released in 2024. In most cases, we did not see an improvement. For the Mistral-Large, Bielik, and GPT-4o models, the latest model even achieves a slightly worse score than earlier versions. The most significant progress can be seen for the Claude Sonnet, especially between the 3.0 and 3.5 versions, but Llama 3 also shows improvement throughout the year.

\section{Conclusions}
In this publication, we introduced the Polish linguistic and cultural competency benchmark, aimed at assessing the ability of LLMs to understand the Polish cultural context. To the best of our knowledge, this is the first evaluation of linguistic and cultural aspects in LLMs for the Polish language. The results of our experiments showed that commercial models still maintain a significant advantage over open-weight models. However, we observe consistent development in both groups, suggesting that multilingual capabilities may significantly improve in the coming months. Furthermore, we demonstrated that training models focused solely on Polish is meaningful, as even models with a small number of parameters such as Bielik, can match much larger multilingual models in understanding cultural aspects. Our benchmark proved to be challenging for the modern LLMs, with the best of them answering about 80\% of the questions correctly. This indicates that there is still considerable room for improvement, and our dataset can continue to be used in the future to track the progress of language models in understanding Polish.

\bibliographystyle{splncsnat}
\bibliography{references}

\end{document}